\documentclass[letterpaper]{article}
\usepackage{uai2020}
\usepackage[margin=1in]{geometry}
\usepackage{microtype}
\usepackage{graphicx}
\usepackage{amsmath}
\usepackage{color}
\usepackage{bbm}
\usepackage{bm}
\usepackage{caption}
\usepackage{amssymb}
\usepackage{mathtools}
\usepackage{subfigure}
\usepackage{algorithmic,algorithm}
\usepackage{multirow}
\usepackage{booktabs} 
\usepackage{amssymb}
\usepackage{url}
\usepackage{tikz}
\usepackage{cite}
\usepackage{bbold}
\usepackage[export]{adjustbox}
\usepackage{apalike}
\newcommand{\mg}{\mathcal{G}}
\newcommand{\md}{\mathcal{D}}

\newcommand{\BX}{\mathbf{X}}
\newcommand{\BZ}{\mathbf{Z}}

\newcommand{\Hquad}{\hspace{0.5em}} 
\DeclareMathOperator*{\argmin}{arg\,min}
\DeclareMathOperator*{\argmax}{arg\,max}

\usepackage{times}

\title{Non-Parametric Graph Learning for Bayesian Graph Neural Networks}

\author{Soumyasundar Pal$^{1*\dagger}$, Saber Malekmohammadi $^2$, Florence Regol$^{1\dagger}$, Yingxue Zhang$^2$, Yishi Xu$^{3\dagger}$, Mark Coates$^1$\\
1. Department of Electrical and Computer Eningeering, McGill University, Montr{\'e}al, QC, Canada\\ 
\{soumyasundar.pal, florence.robert-regol\}@mail.mcgill.ca, mark.coates@mcgill.ca\\
2. Huawei Noah's Ark Lab, Montr{\'e}al Research Center, Montr{\'e}al, QC, Canada \\
\{saber.malekmohammadi, yingxue.zhang\}@huawei.com\\
3. Mila, Universit{\'e} de Montr{\'e}al, Montr{\'e}al, QC, Canada\\
yishi.xu@umontreal.ca
}

\begin{document}

\maketitle
\begin{abstract}
Graphs are ubiquitous in modelling relational structures. Recent
  endeavours in machine learning for graph structured data have led to
  many architectures and learning algorithms. However,  the graph used
  by these algorithms is often constructed based on inaccurate modelling assumptions and/or
  noisy data. As a result, it fails to represent the true
  relationships between nodes. A Bayesian framework which targets
  posterior inference of the graph by considering it as a random
  quantity can be beneficial. In this paper, we propose a novel
  non-parametric graph model for constructing the posterior
  distribution of graph adjacency matrices. The proposed model is flexible
  in the sense that it can effectively take into account the output of
  graph based learning algorithms that target specific tasks.  In addition, model inference 
  scales well to large graphs.  We demonstrate the advantages of this
  model in three different problem settings: node classification, link
  prediction and recommendation.
\end{abstract}
\let\thefootnote\relax\footnotetext{$^*$Corresponding author}
\let\thefootnote\relax\footnotetext{$\dagger$Work done as intern at Huawei Noah's Ark Lab, Montreal Reasearch Center.}

\section{INTRODUCTION}
Growing interest in inference tasks involving networks has prompted
the need for learning architectures adapted to graph-structured
data. As a result, numerous models have been proposed for addressing
various graph based learning tasks such as classification, link
prediction, and recommendation. These approaches process the observed
graph as if it depicts the true relationship among the nodes. In
practice, the observed graphs are formed based on imperfect
observations and incorrect modelling assumptions. Spurious edges might
be formed and important links might be deleted. The vast majority of existing
algorithms cannot take the uncertainty of the graph structure into account during training as there is no mechanism for removing spurious edges and/or adding informative edges in the observed graph. 

Several algorithms that do address this uncertainty by incorporating
a graph learning component have been proposed
recently~\cite{zhang2019,ma2019,tiao2019,jiang2019}. These methods
have limitations, either involving parametric graph models that restrict their
applicability or being focused on the task of node classification. 

In this work, we propose a non-parametric graph inference technique which is incorporated in a Bayesian framework to tackle node and/or edge level learning tasks. Our approach has the following key benefits. First, it generalizes the applicability of the Bayesian techniques outside the realm of parametric modelling. Second, flexible, task specific graph learning can be achieved; this makes effective use of the outputs of existing graph-learning techniques to improve upon them. Third, the graph learning procedure scales well to large graphs, in
contrast to the increased difficulty of parametric approaches.

We conduct extensive experiments to demonstrate the usefulness of our
model for three different graph related tasks. In a node classification setting we observed increased accuracy for
settings where the amount of labeled data is very limited. For the setting of unsupervised learning, we show that incorporating a graph
learning step when performing variational modelling of the graph structure with auto-encoder models leads to better link prediction. Finally, a Bayesian approach based on our proposed model improves recall for existing state-of-the-art graph-based recommender system architectures.

\section{RELATED WORK}

\paragraph{Topology uncertainty in graph neural networks:}

The most closely related work to our proposed approach is a group of recent techniques
that jointly perform inference of the graph while addressing a learning
task such as node classification. The recently proposed Bayesian
GCN~\cite{zhang2019} provides a general, principled framework to deal
with the issue of uncertainty on graphs. Similar ideas are considered
in~\cite{ma2019}, where variational inference is used to learn the
graph structure. This formulation allows consideration of additional
data such as features and labels when performing graph inference, but
the technique is still tied to a parametric model. In~\cite{tiao2019},
the authors take a non-parametric approach, but their probabilistic
formulation is focused on improving only very noisy graphs.
In~\cite{jiang2019}, simultaneous optimization of the graph structure
along with the learning task is considered. In all of these works,
only the node classification task has been explored. Our methodology
extends the applicability of these methods by combining the Bayesian
framework with a more flexible non-parametric graph model.

\paragraph{Graph learning:} Multiple algorithms have been proposed that focus exclusively on
learning graph connectivity based on observed
data~\cite{dong2016,kalofolias2016}. These works differ from
ours in that the end goal is topology inference. These algorithms
typically appeal to a smoothness criterion for the graph. Although
these methods provide useful graphs, they have $\mathcal{O}(N^2)$
complexity. As a result, many do not scale well to
large graphs. Approximate nearest neighbour (A-NN) graph
learning~\cite{malkov2020} has $\mathcal{O}(N \log N)$
complexity, which is more suitable for large scale applications,
but the learned graph generally has poor quality compared to the k-NN
graph. A more recent method in~\cite{kalofolias2019} introduces an
approximate graph learning algorithm which provides an efficient trade
off between runtime and the quality of the solution. We build on this
method for our inference procedure, but our graph model is tailored to the specific learning task we address.

\paragraph{Deep learning based graph generative models:} There is a large body of existing work for deep learning based graph generative models.
 In~\cite{li2018c,simonovsky2018,you2018,liao2019,liu2019} various algorithms for graph generation using VAEs, RNNs, and normalizing flow are developed. These approaches are evaluated based on the likelihood of sampled graphs and comparing graph characteristics. Moreover these algorithms do not preserve node identities, so sampled (inferred) graphs cannot be directly used for node or edge level inference. Generative adversarial networks (GANs) based approaches~\cite{wang2017,bojchevski2018} are more successful in sampling graphs similar to the observed one. However, these models have prohibitively high computational complexity and their performance is heavily dependent on hyperparameter tuning.
 
\paragraph{Node classification:} A central learning task on graphs is semi-supervised node classification. In general, the most common approach is to incorporate graph filters within deep learning
algorithms. Early works~\cite{duvenaud2015,defferrard2016} based their models on theory from the graph signal
processing community. This approach led to more sophisticated graph
convolution architectures~\cite{kipf2017,velivckovic2018,hamilton2017b}. More recent models include~\cite{zhuang18,wijesinghe2019}. In~\cite{tian2019}, a learnable graph kernel based on a data-driven similarity metric is considered for node classification. Our graph learning framework can be combined with these algorithms to augment performance, particularly when there is a very limited amount of labelled data.

\paragraph{Link prediction:}

 Several algorithms based on autoencoders have been shown to perform extremely well for the link prediction task~\cite{kipf2016,pan2018,grover2019,metha2019}. These techniques learn node embeddings in a (variational) autoencoder framework and model the probability of the existence of an edge based on the closeness of the embeddings. We show how our method can be combined with these strategies to deliver a small but consistent improvement for the link prediction task.

\paragraph{Recommender systems:}

Recommender systems have become a key factor to meet users' diverse
and personalized needs for online consumption platforms.  The most
common approach is collaborative filtering (CF).  Recent works have
incorporated graphs and GNNs to better model the user-item interactions~\cite{vandenberg2018,ying2018,wang2019,sun2019,monti2017geometric,zheng2018spectral}.

Although the GNN-based recommendation models have achieved
impressive performance, existing methods regard the provided user–item interaction records as ground truth. In many practical settings,
the user-item interaction graph has spurious edges due to
noisy information; on the other hand, some potential user-item
positive interactions are missing because the item is never presented
to the user. This is falsely indicated as a negative interaction. Thus, it is important to capture the uncertainty in the
observed user-item interaction graph. In the following methodology section, we elaborate on how our graph learning approach can alleviate this problem.

\section{METHODOLOGY}
\subsection{NON-PARAMETRIC GRAPH LEARNING}
\label{sec:non_param}
In many learning tasks, often an observed graph $\mg_{obs}$ provides additional structure to the given data $\md$. The data $\md$ can include feature vectors, labels, and other information, depending on the task at hand. If $\mg_{obs}$ is not readily available, it is often built from the data $\md$ and possibly other side-information.  In many cases, $\mg_{obs}$ does not represent the true relationship of the nodes as it is often formed using inaccurate modelling assumptions and/or is constructed from noisy data. In several recent works~\cite{zhang2019,ma2019,tiao2019}, it has been shown that building a posterior model for the `true' graph $\mg$ and incorporating it in the learning task is beneficial.

We propose a non-parametric generative model for the adjacency matrix $\mathbf{A}_{\mg}$ of the random undirected graph $\mg$. $\mathbf{A}_{\mg}$ is assumed to be a symmetric matrix with non-negative entries. We emphasize that our model retains the identities of the nodes and disallows permutations of nodes (permutations of adjacency matrices are not equivalent graphs when node identities are preserved). This characteristic is essential for its use in node and edge level inference tasks. We define the prior distribution for $\mg$ as
\begin{align}
  p(\mg) \propto
  \begin{dcases}
    e^{\left(\alpha\mathbf{1}^{\top}\log(\mathbf{A}_{\mg}\mathbf{1}) - \beta \|\mathbf{A}_{\mg} \|_F^2\right)}\,, &\text{ if }
      \mathbf{A}_{\mg} \geq \mathbf{0} \\
      &\text{\phantom{ if }} \mathbf{A}_{\mg} = \mathbf{A}_{\mg}^{\top}\\
    0\,, &  \text{otherwise}\,.
    \end{dcases}
\end{align}
The first term in the log prior is a logarithmic barrier on the degree of the nodes which prevents any isolated node in
$\mg$. The second term is a regularizer based on the Frobenius norm which encourages low weights for the links. $\alpha$ and $\beta$ are hyperparameters which control the
scale and sparsity of $A_{\mg}$. In our model, the joint likelihood of $\mg_{obs}$  and $\md$ conditioned on $\mg$ is:
 \begin{align}
    p(\mg_{obs}, \md|\mg) \propto \exp{(- \|\mathbf{A}_{\mg} \circ \mathbf{D}(\mg_{obs}, \md) \|_{1,1})}\,,\label{eq:joint_likelihood}
\end{align}
where
$\mathbf{D}(\mg_{obs}, \md) \geq \mathbf{0}$ is
a symmetric pairwise distance matrix which encodes the dissimilarity between the nodes. The symbol
$\circ$ denotes the Hadamard product and $\|\cdot\|_{1,1}$ denotes the elementwise $\ell_1$ norm. The likelihood encourages higher edge weights for the node pairs with lower pairwise distances and vice versa.

Bayesian inference of the graph $\mg$ involves sampling from its posterior distribution.
The space is high dimensional ($\mathcal{O}(N^2)$, where $N$ is the number of the nodes). Designing a suitable sampling scheme (e.g., Markov Chain Monte Carlo) in such a high dimensional space is extremely challenging and computationally demanding for large graphs. Instead we pursue maximum a posteriori estimation, which is equivalent to approximating the posterior  by a point mass at the mode~\cite{mackay1996}. We solve the following optimization problem:
 \begin{align}
 \widehat{\mg} &= \argmax_{\mg} p(\mg|\mg_{obs}, \md)\,,\label{opt:graph_inference}
 \end{align}
which is equivalent to learning an $N\!\times\!N$ symmetric adjacency matrix of $\widehat{\mg}$.
\begin{align}
      \label{opt:adj_inference}
  \mathbf{A}_{\widehat{\mg}}&= \argmin_{\substack{\mathbf{A}_{\mg} \in \mathbf{R_+}^{N\times N}, \\ \mathbf{A}_{\mg}=\mathbf{A}_{\mg}^{\top} }}
  \begin{aligned}[t]
  &\|\mathbf{A}_{\mg} \circ \mathbf{D}\|_{1,1} -\alpha\mathbf{1}^{\top}\log(\mathbf{A}_{\mg}\mathbf{1}) \\
  & \quad \quad \quad \quad \quad \quad \quad + \beta\|\mathbf{A}_{\mg}\|_F^2 \,.
 \end{aligned}
 \end{align}
  The optimization problem in~\eqref{opt:adj_inference} has been studied
 in the context of graph learning from smooth
 signals. \cite{kalofolias2016} adopts a primal-dual optimization
 technique to solve this problem. However the complexity
 of this approach scales as $\mathcal{O}(N^2)$, which can be
 prohibitive for large graphs. In this paper, we employ the scalable, approximate algorithm in
 ~\cite{kalofolias2019}, which has several advantages as follows. First, it can use existing approximate nearest neighbour techniques, as in~\cite{malkov2020}, to reduce the dimensionality of the optimization problem. Second, the graph learning has a computational complexity of  $\mathcal{O}(N\log N)$ (the same as approximate nearest neighbour algorithms), while the quality of the learned graph is comparable to the state-of-the-art. Third, if we are not concerned about the scale of the learned graph (which is typical in many learning tasks we consider, since a normalized version of the adjacency or Laplacian matrix is used), the approximate algorithm allows us to
 effectively use only one hyperparameter instead of $\alpha$ and $\beta$ to control the sparsity of the solution and provides a useful heuristic for automatically selecting a suitable value based on the desired edge density of the solution.

 In our work, we use this approximate algorithm for inference of the graph $\mg$, which is subsequently used in various learning tasks. Since, we have freedom in choosing a functional form for $\textbf{D}(\cdot, \cdot)$, we can design suitable distance metrics in a task specific manner. This flexibility allows us to incorporate the graph learning step in diverse tasks. In the next three subsections, we present how the graph learning step can be applied to develop Bayesian algorithms for node classification, link prediction and recommendation systems.

\subsection{NODE CLASSIFICATION}
\label{sec:classification}
\paragraph{Problem Statement:}
We consider a semi-supervised node classification problem for the nodes in $\mg_{obs}$. In this setting we also have access to the node attributes $\BX$ and the labels in the training set $\mathbf{Y_{\mathcal{L}}}$. So, $\md = (\BX, \mathbf{Y_{\mathcal{L}}})$. The task is to predict the labels of the remaining nodes $\mathbf{Y_{\overline{\mathcal{L}}}}$, where $\overline{\mathcal{L}} = \mathcal{V} \setminus \mathcal{L}$.

\paragraph{Bayesian GCN -- non-parametric model:}
\cite{zhang2019} derive a Bayesian learning
methodology for GCNs by building a posterior model for $\mg$. Their
approach assumes that $\mg_{obs}$ is sampled from a parametric graph
model. The graph model parameters are marginalized to target inference
of the graph posterior $p(\mg|\mg_{obs})$. Although this approach is
effective, it has several drawbacks. The methodology lacks flexibility
since a particular parametric model might not fit different types
of graph. Bayesian inference of the model parameters is often
challenging for large graphs. Finally, parametric modelling of graphs
cannot use the information provided by the node features $\BX$ and
training labels $\mathbf{Y_{\mathcal{L}}}$ for inference of $\mg$.
Here, we propose to incorporate a non-parametric model for inference
of $\mg$ in the BGCN framework. We aim to compute the marginal
posterior probability of the node labels, which is obtained via
marginalization with respect to the graph $\mg$ and GCN weights $\mathbf{W}$:
\begin{align}
p(\BZ|\mathbf{Y_{\mathcal{L}}},\BX,\mg_{obs}) &= \int p(\BZ|\mathbf{W},\mg_{obs},\BX)p(\mathbf{W}|\mathbf{Y_{\mathcal{L}}},\BX,\mg)\,\nonumber\\
&\qquad p(\mg|\mg_{obs}, \BX, \mathbf{Y_{\mathcal{L}}}) \,d\mathbf{W}\,d\mg \,. \label{eq:exact_posterior}
\end{align}
The categorical distribution of the node labels $p(\BZ|\mathbf{Y_{\mathcal{L}}},\BX,\mg_{obs})$ is modelled by applying a softmax function to the output of the last layer of the GCN. The integral in~\eqref{eq:exact_posterior} cannot be computed in a closed form, so we employ Monte Carlo to approximate it as follows:
\begin{align}
p(\BZ|\mathbf{Y_{\mathcal{L}}},\BX,\mg_{obs}) \approx 
\dfrac{1}{S} \sum_{s=1}^S
p(\BZ|\mathbf{W}_{s},\mg_{obs},\BX)\,.
\label{eq:MC_posterior}
\end{align}
Here, we learn the maximum a posteriori (MAP) estimate $\widehat{\mg} = \displaystyle{\argmax_{\mg}}\Hquad p(\mg|\mg_{obs}, \BX, \mathbf{Y_{\mathcal{L}}})$ and subsequently sample $S$ weight matrices $\mathbf{W}_{s}$ from $p(\mathbf{W}|\mathbf{Y_{\mathcal{L}}},\BX,\widehat{\mg})$ by training a Bayesian GCN using the graph $\widehat{\mg}$.

In order to perform the graph learning step, we need to define a
pairwise distance matrix $\mathbf{D}$. For this application, we
propose to combine the output of a node embedding algorithm and a base
classifier to form $\mathbf{D}$:
\begin{align}
  \mathbf{D}(\BX,\mathbf{Y_{\mathcal{L}}},\mg_{obs}) = \mathbf{D}_{1} (\BX,\mg_{obs}) + \delta \mathbf{D}_{2}(\BX,\mathbf{Y_{\mathcal{L}}},\mg_{obs})\,.\label{eq:distance}   
\end{align}
Here $\delta$ is a hyperparameter which controls the importance of $\mathbf{D}_2$ relative to $\mathbf{D}_1$.
The $(i,j)$'th entries of $\mathbf{D}_1$ and $\mathbf{D}_2$ are defined as follows: 
 \begin{align}
     D_{1,ij} (\BX,\mg_{obs}) &= \|\boldsymbol{z}_i - \boldsymbol{z}_j\|^2\,,\label{eq:dist1}\\
     D_{2,ij}(\BX,\mathbf{Y_{\mathcal{L}}},\mg_{obs}) &= \frac{1}{|\mathcal{N}_i||\mathcal{N}_j|}\displaystyle{\sum_{k \in \mathcal{N}_i}\sum_{l \in \mathcal{N}_j}}\mathbb{1}_{(\hat{c}_k \neq \hat{c}_l)}\,.\label{eq:dist2}
 \end{align}
 Here, $\boldsymbol{z}_i$ is any suitable embedding of node $i$ and $\hat{c}_i$
 is the predicted label at node $i$ obtained from the base classification algorithm. $\mathbf{D}_1$ measures pairwise dissimilarity in terms of the observed topology
 and features and $\mathbf{D}_2$ summarizes the discrepancy of the node labels in the neighbourhood. For the experiments, we choose the Variational
 Graph Auto-Encoder (VGAE) algorithm~\cite{kipf2016} as the node embedding
 method to obtain the $\boldsymbol{z}_i$ vectors and use the GCN proposed by~\cite{kipf2017} as the base
 classifier to obtain the $\hat{c}_i$ values. The neighbourhood of the $i$-th node is defined as:
 \begin{align}
 \mathcal{N}_i = \{j | (i,j) \in \mathcal{E}_{\mg_{obs}}\} \cup\{i\}\,.\nonumber
 \end{align}
Here, $\mathcal{E}_{\mg_{obs}}$ is the set of edges in $\mg_{obs}$. With the regard to the choice of the hyperparameter $\delta$, we
observe that
 \begin{align}
 \delta = \dfrac{\displaystyle{\max_{i,j}} \Hquad D_{1,ij}}{\displaystyle{\max_{i,j}} \Hquad D_{2,ij}}\nonumber
 \end{align}
works well in our experiments, although it can be tuned via cross-validation if a validation set is available.

For the inference of GCN weights $\mathbf{W}$, many existing algorithms such as expectation propagation~\cite{hernandez2015}, variational inference~\cite{gal2016,sun2017}, and Markov Chain Monte Carlo methods~\cite{neal1993,li2016d} can be employed. As in~\cite{zhang2019}, we train a GCN on the inferred graph $\widehat{\mg}$ and use Monte Carlo dropout~\cite{gal2016}. This is equivalent to sampling $\mathbf{W}_s$ from a particular variational approximation of $p(\mathbf{W}|\mathbf{Y_{\mathcal{L}}},\BX,\widehat{\mg})$. The resulting algorithm is provided in the supplementary material.

\subsection{LINK PREDICTION}
\paragraph{Problem statement:}
In this setting, some of the links in $\mg_{obs}$ are hidden or unobserved. The task is to predict the unseen links based on the knowledge of the (partially) observed $\mg_{obs}$ and the node features $\BX$. Thus in this case, the additional data beyond the graph is $\md = \BX$. 

\paragraph{Background:}
In existing works, the link prediction problem is addressed by
building deep learning based generative models for graphs. In
particular, various architectures of graph variational
auto-encoders~\cite{kipf2016,grover2019,metha2019} aim to learn the posterior distribution of the node embedding $\boldsymbol{Z}$
conditioned on the observed graph $\mg_{obs}$ and the node features $\BX$. The inference model (encoder) often uses simplifying assumptions (e.g. mean-field approximation over nodes or diagonal covariance structures) for the parametric form of the approximate variational posterior distribution $q(\boldsymbol{Z}|\mg_{obs}, \BX)$.
Deep learning architectures are used to learn the parameters of the model. The decoder is another deep learning model which explains how the graph is generated from the embeddings, i.e., it parameterizes $p(\mg_{obs}|\boldsymbol{Z}, \BX)$. Typically the probability of a link in these models is dependent on the similarity of the embedding of the two incident nodes. Assuming a suitable prior $p(\boldsymbol{Z})$, the encoder and decoder is trained jointly to minimize the KL divergence between $q(\boldsymbol{Z}|\mg_{obs}, \BX)$ and the true posterior
$p(\boldsymbol{Z}|\mg_{obs}, \BX)$. The learned embeddings are
evaluated based on an amortized link prediction task for the unseen portion of the graph.

\paragraph{Proposed methodology -- Bayesian VGAE:}
We consider a Bayesian formulation, where we conduct Bayesian
inference of the graph $\mg$ in the encoder. Let us introduce a
function $\mathcal{J}(\mg, \mg_{obs})$ that returns a graph such that
the unobserved entries of the adjacency matrix of $\mg_{obs}$ are
replaced by the corresponding entries of $\mg$.  We
then model the inference distribution as follows:
\begin{align}
q(\boldsymbol{Z}|\mg_{obs}, \BX) &= \int q(\boldsymbol{Z}|\mathcal{J}(\mg, \mg_{obs}), \BX) p(\mg|\mg_{obs}, \BX) d \mg\,,\nonumber\\
&\approx q(\boldsymbol{Z}|\mathcal{J}(\widehat{\mg}, \mg_{obs}), \BX)\,,\nonumber
\end{align}
where
$\widehat{\mg} = \displaystyle{\argmax_{\mg}} \Hquad p(\mg|\mg_{obs}, \BX)$
is the MAP estimate from the non-parametric model. The intuitive idea
behind this modeling is that if the non-parametric inference provides
a reasonable approximation of the unobserved adjacency matrix entries, then an auto encoder
trained on a graph that incorporates these approximate entries should learn better embeddings.
For the graph learning step, we form the distance matrix $\mathbf{D}$ using the output of an
auto-encoder as follows:
\begin{align}
     D_{ij} (\BX,\mg_{obs}) &= \|\mathbb{E}_q[\boldsymbol{z}_i] - \mathbb{E}_q[\boldsymbol{z}_j]\|^2\,.\label{eq:dist_gae}
\end{align}
The resulting algorithm is summarized in the supplementary material.

\subsection{RECOMMENDATION SYSTEMS}
\paragraph {Problem statement:}
In this section we address a personalized item recommendation task
based on historical interaction data. We denote the set of users and
items by $\mathcal{U}$ and $\mathcal{I}$ respectively. The
interaction between any user $u \in \mathcal{U}$ and item
$i \in \mathcal{I}$ is encoded as a link in a bipartite graph
$\mg_{obs}$. The task is to infer the unobserved interactions (and to
use these as predictions of future interactions). Viewed in this
light, the recommendation task is a link prediction problem. However, in many
cases, predicting a personalized ranking for the items is
important~\cite{rendle2009}.

For each user $u$, if there is an observed interaction with item $i$
and an unobserved interaction with item $j$, we write that $i >_u j$
in the training set. The introduced relation $i >_u j$ implies that
user $u$ prefers item $i$ to item $j$. This interaction training data
leads to a set of rankings $\{>_u\}_{trn}$ for each user $u$ over the
training set of triples: $\{(u,i,j) : (u,i) \in \mg_{obs}, (u,j) \notin
\mg_{obs}\}$. We denote these rankings for all users in $\mathcal{U}$
as $\{>_{\mathcal{U}}\}_{trn}$. This training data is used to learn a model parameterized by
$\mathbf{W}$. The generalization capability is tested by ranking, for
each user $u$, all
$(u,i, j)$ such that both $(u,i)$ and $(u,j) \notin \mg_{obs}$. We
denote the rankings for a specific user in this test set $\{(u,i,j) : (u,i) \notin \mg_{obs}, (u,j) \notin
\mg_{obs}\}$ as $\{>_u\}_{test}$. The collection of all such rankings
for all users is denoted $\{>_{\mathcal{U}}\}_{test}$
In this paper, we propose to incorporate Bayesian
inference of graph $\mg$ in the Bayesian Personalized Ranking
(BPR) loss formulation~\cite{rendle2009}. A brief review of the BPR loss is provided for completeness.

\paragraph{Background -- BPR loss:}
Many existing graph based deep learning recommender
systems~\cite{sun2019,wang2019,ying2018}
learn an embedding $e_u(\mathbf{W}, \mathcal{G}_{obs})$ for user $u$
and $e_i(\mathbf{W}, \mathcal{G}_{obs})$ for item $i$ and model the
probability that user $u$ prefers item $i$ to item $j$ as follows:
\begin{align}
p(i >_u j|\mathcal{G}_{obs}, \mathbf{W}) = \sigma(e_u \boldsymbol{\cdot} e_i - e_u \boldsymbol{\cdot} e_j)\,.\nonumber
\end{align}
Here $\sigma(\cdot)$ is the sigmoid function and $\boldsymbol{\cdot}$ is the inner product. Our goal is to compute: 
\begin{align}
p(\{>_{\mathcal{U}}\}_{test}|&\{>_{\mathcal{U}}\}_{train}, \mg_{obs}) = \int p(\{>_{\mathcal{U}}\}_{test}|\mg_{obs}, \mathbf{W})\,\nonumber\\ &p(\mathbf{W}|\{>_{\mathcal{U}}\}_{train},\mg_{obs}) d\mathbf{W}\,,
\end{align}
but this integral is not tractable. In practice, we assume a prior
$\mathcal{N}(\mathbf{0}, \lambda^{-1} I)$ for $\mathbf{W}$ and model
the preferences of different users as independent. We can then consider a MAP
estimate of $\mathbf{W}$:
\begin{align}
\widehat{\mathbf{W}} &= \argmax_{\mathbf{W}} p(\mathbf{W}|\{>_{\mathcal{U}}\}_{train},\mg_{obs}) \,,\nonumber\\
&= \argmax_{\mathbf{W}} p(\mathbf{W})p(\{>_{\mathcal{U}}\}_{train},\mg_{obs} | \mathbf{W})\,,\nonumber \\
&= \argmax_{\mathbf{W}}  \bigg( -\frac{\lambda} {2} \lvert \lvert \mathbf{W} \rvert \rvert^2 +\,\nonumber\\
&\quad \quad \sum_{(u,i,j) \in \{>_{\mathcal{U}}\}_{trn}} \log \left(\sigma(e_u \boldsymbol{\cdot} e_i -e_u \boldsymbol{\cdot}\cdot e_j\right)\bigg)\,.\nonumber
\end{align}
This is equivalent to minimizing the BPR loss, where the positive pool
$\{(u,i) : (u,i) \in \mg_{obs}\}$ and negative pool
$\{(u,j) : (u,j) \notin \mg_{obs}\}$ are created according to
$\mg_{obs}$. Once the MAP estimate has been obtained, we assess the performance by ranking the test set
triples using $\widehat{\mathbf{W}}$.

\paragraph{Non-parametric model -- Bayesian graph recommender system:}

In the Bayesian setting, ranking is conducted by considering an expectation with respect to the posterior distribution of the graph $\mg$ from the
non-parametric model $p(\mg|\mg_{obs}, \{>_{\mathcal{U}}\}_{train})$. We need to evaluate the posterior
probability of ranking in the test set. Let us introduce the graph
$\widetilde{G} = \mathcal{J}_r(\mg, \mg_{obs})$, which is obtained via a
function $\mathcal{J}_r$ that combines the information in $\mg$ and
$\mg_{obs}$. We specify the function $\mathcal{J}_r$ that we employ in
our methodology more precisely below. We can then write the posterior
probability of the ranking of the test set as follows:
\begin{align}
&p(\{>_{\mathcal{U}}\}_{test}|\{>_{\mathcal{U}}\}_{train},\mg_{obs}) = \int p(\{>_{\mathcal{U}}\}_{test}|\mg_{obs}, \mathbf{W})\,\nonumber\\
& \quad p( \mathbf{W}|\{>_{\mathcal{U}}\}_{train}, \widetilde{\mg}) p(\mg| \mg_{obs}, \{>_{\mathcal{U}}\}_{train})\,d\mg\,d \mathbf{W} \,. \label{eq:bbpr}
\end{align}
We approximate the integrals with respect to the posteriors of $\mg$ and $\mathbf{W}$
by the MAP estimates to obtain:
\begin{align}
&p(\{>_{\mathcal{U}}\}_{test}|\{>_{\mathcal{U}}\}_{train},\mg_{obs}) \approx p(\{>_{\mathcal{U}}\}_{test}|\mg_{obs}, \widehat{\mathbf{W}})\,.\label{eq:mc"_bbpr}
\end{align}
To calculate this approximation we first perform the non-parametric
graph learning to obtain
$\widehat{\mg} = \displaystyle{\argmax_{\mg}}\Hquad p(\mg| \mg_{obs},
\{>_u\}_{train})$, then compute the new graph
$\widetilde{G} = \mathcal{J}_r(\widehat{\mg}, \mg_{obs})$ and minimize the BPR loss to
form the estimate of the weights
\begin{align}
    \widehat{\mathbf{W}} = \argmax_{\mathbf{W}} \Hquad p( \mathbf{W}|\{>_u\}_{train}, \widetilde{G}) 
\end{align}
according to the positive and negative pool defined by this new graph
$\widetilde{G} = \mathcal{J}_r(\widehat{\mg}, \mg_{obs})$.

Since the dot product measures the similarity between the embeddings in the proposed recommender system architecture,  we use the pairwise cosine distance between the learned embedding of a base node embedding algorithm for learning a bipartite graph.
\begin{align}
D_{u,i}(\{>_{\mathcal{U}}\}_{train}, \mg_{obs}) = 1 - \frac{e_u \cdot e_i}{\lvert \lvert e_u \rvert \rvert_2 \lvert \lvert e_i \rvert \rvert_2 }\,.\label{eq:dist_cosine}
\end{align}
Here, the $e_u$'s and $e_i$'s are obtained from the node embedding algorithm. Since in $\mg_{obs}$, none of the test set user-item interactions are present, they are all included in the negative pool. We use the estimated graph $\widehat{\mg}$ to remove potentially positive interactions in the test set from the negative pool. This is achieved by constructing $\mathcal{J}(\widehat{\mg}, \mg_{obs})$ as follows. We identify a fraction of links with the highest edge weights in $\widehat{\mg}$ and subsequently remove them form the negative pool of interactions for the Bayesian approach. The number of links to be removed is decided based on examining the performance on a validation set. The resulting algorithm is summarized in the supplementary material.

\section{EXPERIMENTS}

\subsection{NODE CLASSIFICATION}
\label{sec:classification_results}
We consider a semi-supervised node classification task on three
benchmark citation networks Cora, Citeseer~\cite{sen2008} 
and Pubmed~\cite{namata2012}. 
The details of the datasets are included in the supplementary material. The attribute vector at a node is a sparse bag-of-words
extracted from the keywords in the article and the label denotes the
research topic addressed in the article. We consider three different
experimental settings where we have 5, 10 and 20 labeled nodes per
class in the training set. In each setting, we conduct 50 trials based
on random splitting of the data and random initialization of the
learnable weights. We compare the proposed BGCN with
the ChebyNet~\cite{defferrard2016}, the GCN~\cite{kipf2017},
the GAT~\cite{velivckovic2018}, the DFNET~\cite{wijesinghe2019} (for only Cora
and Citeseer due to runtime considerations), the SBM-GCN~\cite{ma2019} and the BGCN
in~\cite{zhang2019}. The hyperparameters for the GCN are set to those
reported in~\cite{kipf2017} and the same values are used for the BGCNs. We report the average classification
accuracies along with their standard errors in
Table~\ref{table:bgcnresult}. For each setting, we conduct a Wilcoxon signed rank test to determine whether the best performing algorithm is significantly better than the second-best. Results in bold font indicate statistical significance at the 5\% level.
\begin{table}[ht]
\caption{Accuracy of semi-supervised node classification.}
\vspace{0.1cm}
\setlength{\tabcolsep}{5pt}
\centering
\footnotesize{
	\begin{tabular}{llccc}
		\toprule[0.25ex]
		& \textbf{Algorithms} &\textbf{5 labels}        & \textbf{10 labels}         & \textbf{20 labels} \\ 
	\midrule[0.25ex]
	
		\multirow{7}{*}{\rotatebox[origin=c]{90}{\textbf{Cora}}} &\textbf{ChebyNet}             &61.7$\pm$6.8            &72.5$\pm$3.4       &       78.8$\pm$1.6        \\
	&	\textbf{GCN}             &70.0$\pm$3.7            & 76.0$\pm$2.2     &       79.8$\pm$1.8  \\
	&	\textbf{GAT}              &70.4$\pm$3.7            &76.6$\pm$2.8       &      79.9$\pm$1.8   \\
		&	\textbf{DFNET-ATT}              &72.3$\pm$2.9&75.8 $\pm$1.7 &79.3$\pm$1.8  \\
		&\textbf{SBM-GCN}            &   46.0$\pm$19        & 74.4$\pm$10   &    \textbf{82.6$\pm$0.2} \\
	&	\textbf{BGCN}     &74.6$\pm$2.8   &\textbf{77.5$\pm$2.6}  & 80.2$\pm$1.5 \\
	&	\textbf{BGCN (ours)}     &74.2$\pm$2.8   &76.9$\pm$2.2  &   78.8$\pm$1.7 \\
      
		\midrule
		 	\multirow{7}{*}{\rotatebox[origin=c]{90}{\textbf{Citeseer}}}  & \textbf{ChebyNet}             &58.5$\pm$4.8            &65.8$\pm$2.8              &67.5$\pm$1.9             \\
	&	\textbf{GCN}             &58.5$\pm$4.7            &65.4$\pm$2.6              &67.8$\pm$2.3        \\
	&	\textbf{GAT}              &56.7$\pm$5.1            &64.1$\pm$3.3              &67.6$\pm$2.3       \\
		&	\textbf{DFNET-ATT}              & 60.5$\pm$1.2& 63.2 $\pm$2.9 & 66.3$\pm$1.7  \\
		
		& \textbf{SBM-GCN} &    24.5$\pm$7.3    & 43.3$\pm$12  & 66.1$\pm$5.7           \\
		 
	&	\textbf{BGCN}     &63.0$\pm$4.8   &69.9$\pm$2.3     &71.1$\pm$1.8   \\
	&	\textbf{BGCN (ours)}     &\textbf{64.9$\pm$4.6}   &70.1$\pm$1.9     &71.4$\pm$1.6   \\

		\midrule
			\multirow{6}{*}{\rotatebox[origin=c]{90}{\textbf{Pubmed}}} & \textbf{ChebyNet}             &62.7$\pm$6.9            &68.6$\pm$5.0              &74.3$\pm$3.0             \\
	&	\textbf{GCN}              &69.7$\pm$4.5           &73.9$\pm$3.4    &77.5$\pm$2.5  \\
	&	\textbf{GAT}&68.0$\pm$4.8           &72.6$\pm$3.6             &76.4$\pm$3.0\\
	& \textbf{SBM-GCN}            &   59.0$\pm$10        &  67.8$\pm$6.9  &   74.6$\pm$4.5 \\
	&	\textbf{BGCN}     &70.2$\pm$4.5  &73.3$\pm$3.1             &76.0$\pm$2.6  \\
	&	\textbf{BGCN (ours)}     &\textbf{71.1$\pm$4.4}  &\textbf{74.6$\pm$3.6}             &77.6$\pm$2.9  \\
		


		\bottomrule[0.25ex]
	\end{tabular}
}

\label{table:bgcnresult}
\end{table}

\begin{figure}[ht!]
\centering
 \includegraphics[scale=0.45, clip, trim={0.0cm 0  0.0cm 0cm}]
 {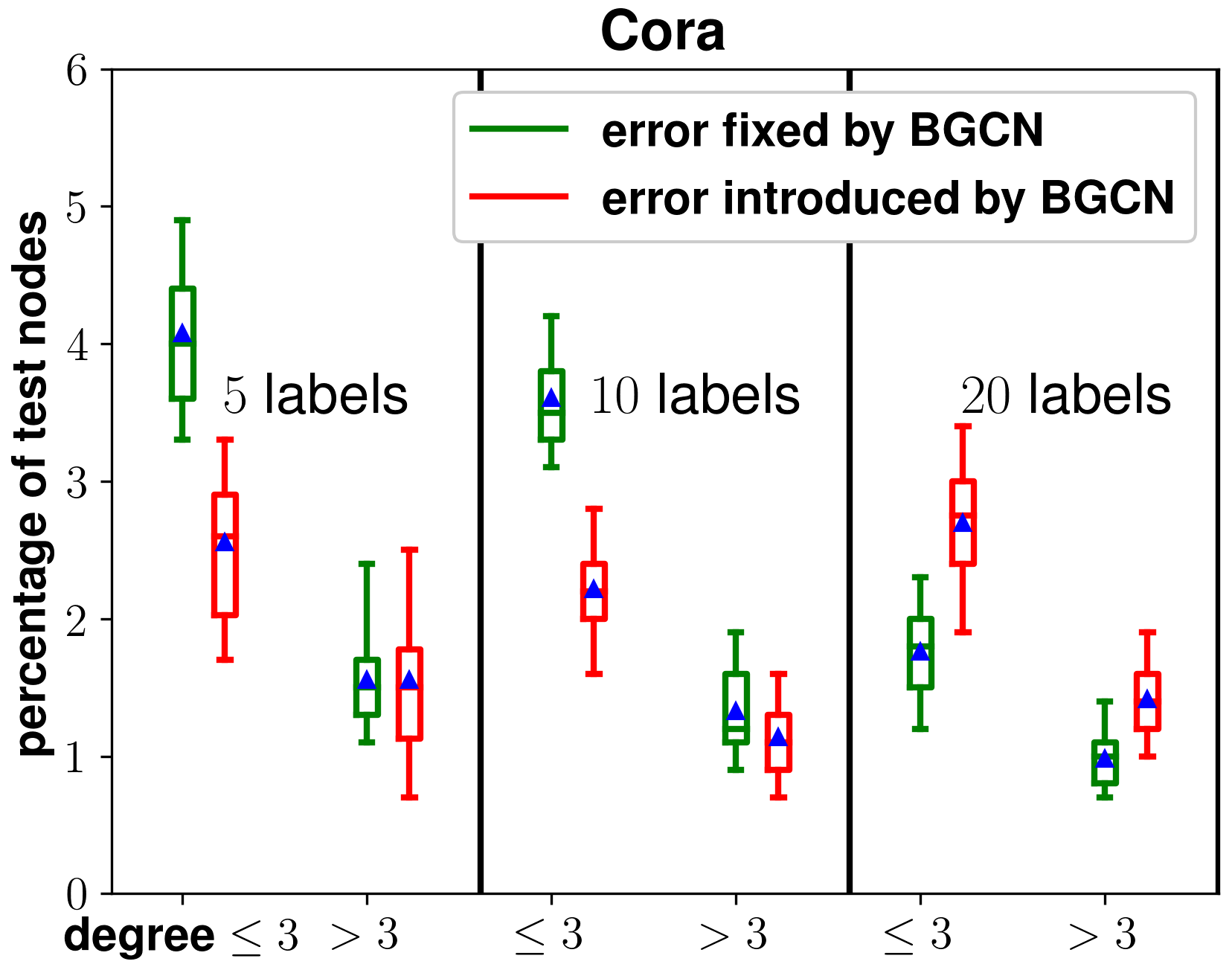}

 \includegraphics[scale=0.45, trim={0.0cm 0 0.0cm 0cm}, clip]
 {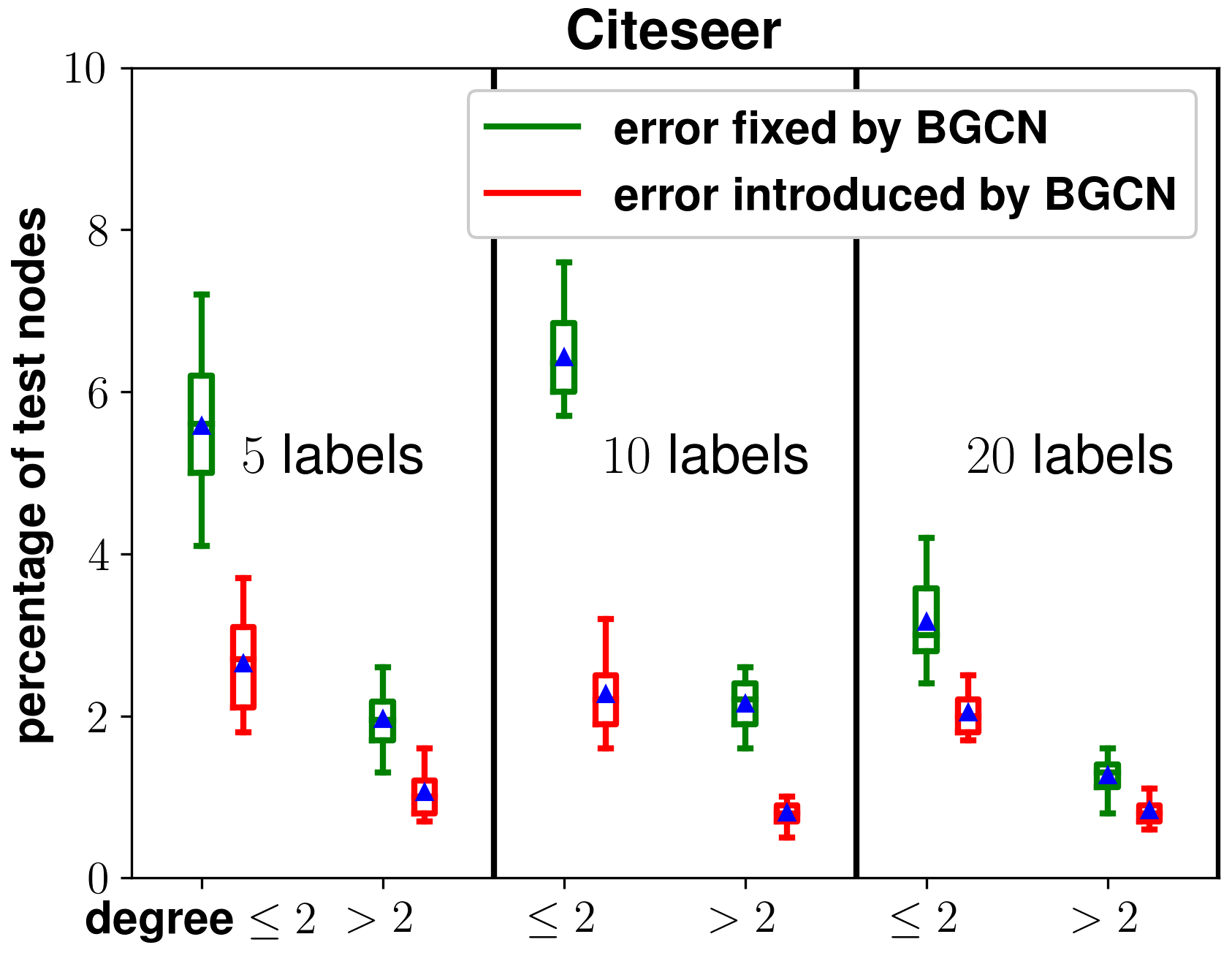}

  \caption{Boxplots of different categories of nodes in the Cora and  Citeseer datasets based on the classification results of the GCN and the proposed BGCN algorithms. The two groups are formed by thresholding the degree of the nodes in the test set at the median value.}
\label{fig:degree_cora}
\end{figure}

\begin{figure}[ht!]
 \includegraphics[scale=0.45, clip, trim={0.9cm 0  0.7cm 1cm},left]
 {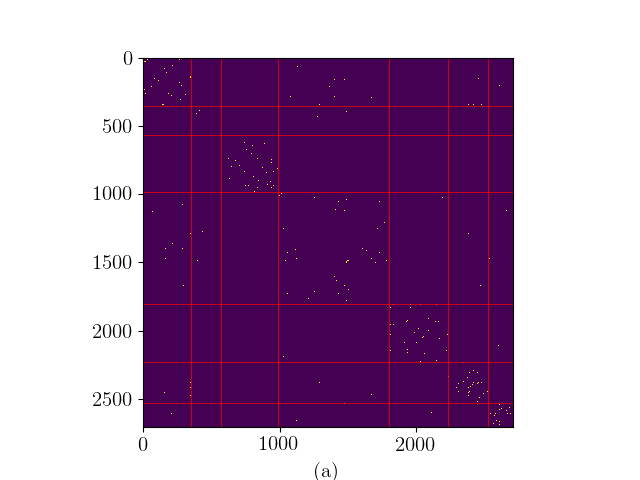}
 \includegraphics[scale=0.45, trim={0.7cm 0 0.7cm 1cm}, clip,left]
 {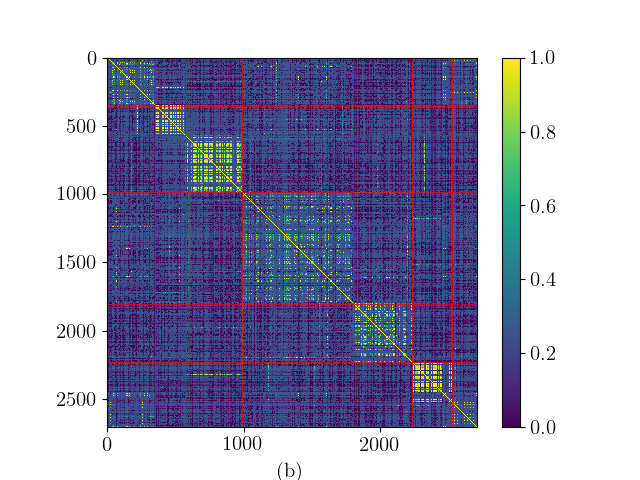}
  \caption{(a) the observed adjacency matrix ($A_{\mg_{obs}})$ and (b) the MAP estimate of adjacency matrix ($A_{\widehat{\mg}}$) from the non-parametric model for the Cora dataset. The node are reordered based on labels. The red lines show the class boundaries.}
\label{fig:adj}
\vspace{-2em}
\end{figure}

The results in Table~\ref{table:bgcnresult} show that the proposed
BGCN with non-parametric modelling of the graph achieves either higher
or competitive accuracies in most cases. The relative improvement
compared to the GCN is more significant if the labelled data is
scarce. Comparison with the BGCN approach based on parametric
modelling in~\cite{zhang2019} demonstrates that better or comparable
accuracies can be achieved from this model, even if we do not target
modelling the community structure of the graph explicitly. From
Figure~\ref{fig:degree_cora}, we observe that in most cases, for the
Cora and the Citeseer datasets, the proposed BGCN algorithm corrects
more misclassifications of the GCN for low degree nodes. The same trend is observed for the Pubmed dataset. The empirical success of the GCN is primarily due to aggregating information with neighbors.  As the low  degree  nodes  have  less opportunity to aggregate, performance is worse at these nodes. The proposed BGCN approach generates many additional links between similar nodes (Fig.~\ref{fig:adj}). This improves learning, particularly at low degree nodes.

In
Figure~\ref{fig:adj}, we compare the adjacency matrix
($A_{\widehat{\mathcal{G}}}$) of the MAP estimate graph
$\widehat{\mg}$ with the observed adjacency matrix $A_{\mg_{obs}}$ for
the Cora dataset. This reveals that compared to $A_{\mg_{obs}}$,
$A_{\widehat{\mg}}$ has denser connectivity among the nodes with the
same label. This provides a rationale of why the proposed BGCN
outperforms the GCN in most cases.

\vspace{-0.25em}

\subsection{LINK PREDICTION}
We consider a link prediction task to demonstrate the usefulness of the learned embeddings from the Bayesian approach. We split the links in 85/5/10\% for training, validation and testing respectively. The validation and test sets contain the same number of non-links as links. During model training, the links in the validation and test sets are hidden while the node features are unaltered. We compare the Bayesian approach with the GAE and VGAE~\cite{kipf2016}, the GRAPHITE-AE and VAE~\cite{grover2019} and the DGLFRM~\cite{metha2019} models. The hyperparameters of these baseline algorithms are selected according to the corresponding papers. Other common baselines, e.g. spectral
Clustering~\cite{tang2011}, Deepwalk~\cite{perozzi2014} and node2vec~\cite{grover2016} are not included since it has been demonstrated that the baselines we include significantly outperform them. We incorporate the non-parametric graph inference technique in the existing auto-encoders to build a Bayesian version of these algorithms. The Area Under the ROC Curve (AUC) and
the Average Precision (AP) score are used as performance
metrics. Table~\ref{table:auc} shows the mean AUC and AP, together with standard errors, based on
50 trials. Each trial corresponds to a random split of the graph
and a random initialization of the learnable parameters. We conduct a Wilcoxon signed rank test to determine the statistical significance of the improvement compared to the corresponding base model. Results in bold font indicate settings where the test declares a significance at the 5\% level.

\begin{table}[ht]
\caption{Area Under the ROC Curve (AUC) and Average Precision (AP) score for link prediction (in \%).}
\vspace{0.1cm}
\setlength{\tabcolsep}{5pt}
\centering
\footnotesize{
\begin{tabular}{lccc}
\toprule[0.25ex]
\textbf{Algorithm}  &\textbf{Cora}        & \textbf{Citeseer}  & \textbf{Pubmed}  \\\midrule[0.25ex]
  & \multicolumn{3}{c}{AUC}   \\  
\midrule[0.25ex]

\textbf{GAE}             &91.5$\pm$0.9            &89.4$\pm$1.5          &  96.2$\pm$0.2  \\
	\textbf{BGAE}             &\textbf{91.8$\pm$0.8}            &\textbf{89.6$\pm$1.6}  &  96.2$\pm$0.2         \\ \hline
	\textbf{VGAE}             &91.8$\pm$0.9            &90.7$\pm$1.0       &94.5$\pm$0.7 \\
	\textbf{BVGAE}             &\textbf{92.2$\pm$0.8}            &\textbf{91.2$\pm$1.0}    & 94.4$\pm$0.7  \\ \hline
	\textbf{Graphite-AE}             &92.0$\pm$0.9            &90.8$\pm$1.1 & 96.0$\pm$0.4 \\
	\textbf{BGraphite-AE}             &\textbf{92.4$\pm$0.9}            &\textbf{91.1$\pm$1.1}  & 96.0$\pm$0.4 \\ \hline
	\textbf{Graphite-VAE}             &92.3$\pm$0.8            &90.9$\pm$1.1     &  95.2$\pm$0.4\\
	\textbf{BGraphite-VAE}             &\textbf{92.7$\pm$0.8}            &\textbf{91.4$\pm$1.1}& 95.2$\pm$0.4    \\  \hline
	\textbf{DGLFRM}             &93.1$\pm$0.6            &93.9$\pm$0.7     &  95.9$\pm$0.1 \\
	\textbf{BDGLFRM}             &\textbf{93.2$\pm$0.6}           &\textbf{94.1$\pm$0.7}      & 95.9$\pm$0.2      \\ 
\midrule[0.25ex]

  & \multicolumn{3}{c}{AP}   \\  
\midrule[0.25ex]

\textbf{GAE}        &    92.6$\pm$0.9           &90.0$\pm$1.7  &96.3$\pm$0.3\\
	\textbf{BGAE}               & \textbf{92.8 $\pm$ 0.9}            &\textbf{90.2$\pm$1.7}  & 96.3$\pm$0.2\\ \hline
	\textbf{VGAE}           &92.9$\pm$0.7            &92.0$\pm$1.0   &  94.7$\pm$0.6\\
	\textbf{BVGAE}           &\textbf{93.3$\pm$0.7}            &\textbf{92.5$\pm$1.0}  & 94.6$\pm$0.6\\ \hline
	\textbf{Graphite-AE}             &92.8$\pm$0.9            &91.6$\pm$1.1     &  96.0$\pm$0.4\\
	\textbf{BGraphite-AE}          &\textbf{93.1$\pm$0.9}            &\textbf{92.0$\pm$1.1}  & 96.0$\pm$0.4\\ \hline
	\textbf{Graphite-VAE}         &93.3$\pm$0.7            &92.1$\pm$1.0     & 95.3$\pm$0.4\\
	\textbf{BGraphite-VAE}           &\textbf{93.7$\pm$0.7}            &\textbf{92.6$\pm$1.0}  & 95.3$\pm$0.4\\  \hline
	\textbf{DGLFRM}           &93.8$\pm$0.6            &94.5$\pm$0.7    & 96.4$\pm$0.1 \\
	\textbf{BDGLFRM}           &\textbf{93.9$\pm$0.6}           &\textbf{94.7$\pm$0.7}    &  96.3$\pm$0.1 \\ 
\bottomrule[0.25ex]
\end{tabular}
\vspace{-2em}
}

\label{table:auc}
\end{table}

From the results in Table~\ref{table:auc}, we observe the proposed approach improves link prediction performance for the Cora and Citeseer datasets compared to the baseline auto-encoder models. The improvement is small but consistent over almost all of the random trials. No improvement is observed for Pubmed. To examine this further, we conducted an experiment where the ground-truth for the test set was provided to the autoencoders. The performance does not change from the reported values; this suggests that the models have reached accuracy limits for the Pubmed dataset.

\subsection{RECOMMENDATION SYSTEMS}

We investigate the performance of the proposed Bayesian method on four real-world and publicly available datasets: ML100K, Amazon-Books, Amazon-CDs and Yelp2018. For each dataset, we conduct pre-processing to ensure that each node in the dataset has sufficient interactions. We consider two threshold values $th_1$ and $th_2$, and filter out those users and those items with fewer than $th_1$ and $th_2$ interactions, respectively. For each user, we split each dataset's existing interaction records into training, validation and test set with the ratio of 70/10/20. We evaluate the model performance using Recall@k and NDCG@k, which are the coverage of true items in the top-k recommendations, and a measure of recommendation ranking quality, respectively. Details of statistics of each dataset after the preprocessing step and the definitions of the evaluation metrics are included in the supplementary material.

\begin{table*}[ht!]
\caption{Recall@10, NDCG@10, Recall@20 and NDCG@10 for the four datasets.}
\label{tab:rec_result}
\centering

\begin{tabular}{lcccclcccc}
\midrule[0.15ex]
\textbf{Amazon-CDs}    & \textbf{R@10}       & \textbf{R@20}     & \textbf{N@10}     & \textbf{N@20}                    & \textbf{Yelp2018}    & \textbf{R@10}       & \textbf{R@20}     & \textbf{N@10}     & \textbf{N@20}         \\ \midrule[0.15ex]
\textbf{MGCCF}         &10.1\%

               & 16.1\%

               & 13.1\%               & 16.9\%                         & \textbf{MGCCF}                           & 7.5\%
               & 12.7\%         & 13.0\%           & 17.4\%                          \\
\textbf{BMGCCF}           & \textbf{10.6}\%          & \textbf{17.0}\% 
               & \textbf{13.4} \%
              & \textbf{17.3}\%                           & \textbf{BMGCCF}                              & {7.6}\%
               & {13.0}\%             & {13.2}\%
           & \textbf{17.7}\%                          \\\midrule[0.15ex]
\textbf{NGCF}         & 8.1\%	
            & 13.5\%               & 11.4\%               & 13.8\%                        & \textbf{NGCF}                           &  6.6\%	
    & 11.3\%              & 11.5\%             & 15.3\%                  \\
\textbf{BNGCF}       & \textbf{9.9}\%
            & 	\textbf{16.2}\%              & \textbf{12.8}\%               & \textbf{16.6}\%                            & \textbf{BNGCF}                          &  {6.7}\%	
             & {11.4}\%              & 11.5\%               & {15.5}\%                        
      \\ \midrule[0.25ex]

\textbf{Amazon-Books}    & \textbf{R@10}       & \textbf{R@20}     & \textbf{N@10}     & \textbf{N@20}                    & \textbf{ML100K}    & \textbf{R@10}       & \textbf{R@20}     & \textbf{N@10}     & \textbf{N@20}         \\ \midrule[0.15ex]
\textbf{MGCCF}         &   10.3 \%          &    16.6\%          &   15.0 \%            &   19.4\%                     & \textbf{MGCCF}                            &  18.3\%
        &     29.4\%         &     25.6\%           &  30.9\%                        \\ 
\textbf{BMGCCF}           &   10.3\%	
           &    16.4\%         &   14.8\%        &  	 19.3\%                       & \textbf{BMGCCF}                             &   {18.4}\%	
           &     {29.5}\%     &       \textbf{25.9}\%    &  	\textbf{31.4}\%                  \\ \midrule[0.15ex]
\textbf{NGCF}         &      8.7\%	
         &        14.5\%      & 13.6\%              & 17.8\%                       & \textbf{NGCF}                           &   17.7\%
            &    {29.0}\%     &   25.3\%             &  30.3\%                  \\
\textbf{BNGCF}       &    \textbf{10.2}\%	
          &    \textbf{16.8}\%           &   \textbf{ 15.2}\%          &       \textbf{19.6}\%                     & \textbf{BNGCF}                          &     {17.7}\%         &  {28.9}\%           & 25.2\%               & 30.1\%                        
      \\ \midrule[0.15ex]


\end{tabular}
\end{table*}

We apply our proposed Bayesian graph-based recommendation formulation
to two recent graph-based recommendation models: the \textbf{MGCCF}
\cite{sun2019} and the \textbf{NGCF}~\cite{wang2019}.

We first train the two algorithms with early stopping patience of 50 epochs to get the embedding vectors for users and items. These are  used to calculate the pairwise cosine distance metrics $\mathbf{D}$ for our proposed graph optimizer. We refer to these original algorithms as ``base models''. We build our proposed models (BMGCCF and BNGCF) on top of the base
models via the following procedure. We first apply edge dropping with a threshold $\tau$ to shrink each dataset's negative edge candidate set. We further train the base models with this optimized negative edge pool with an early stop patience of 100. For a fair comparison, to obtain the baseline performance, we also conduct further training of the models with the original negative edge pool with the same early stop patience setting. We use grid search to determine the percentage of the inferred links with the highest edge weights to be removed from the negative pool. A suitable value is chosen for each dataset from \{1, 2, 5, 10, 20\}\%.


We report the Recall@k and the NDCG@k $(\text{k} = 10, 20)$ of the proposed Bayesian models (BMGCCF and BNGCF) along with those of the base models MGCCF and NGCF for four datasets in Table~\ref{tab:rec_result}. We conduct Wilcoxon signed rank test in each case to determine the significance of the obtained result from the Bayesian model over the corresponding base model. Bold numbers indicate a statistically significant difference at 5\% level between the base model and the Bayesian version of the algorithm.  The advantages of our proposed Bayesian framework can be observed for both base models and across both evaluation metrics. For the much denser ML-100K dataset, the procedure is less effective (and in some cases ineffective). With many more edges in the observed graph, the graph-based recommender system algorithms already have considerable information. Although the inferred graph does remove many incorrect edges from the negative pool, this has only a minor impact on the learned embeddings.   

\begin{figure}[ht!]
  \centering
  \includegraphics[width=0.9\linewidth]{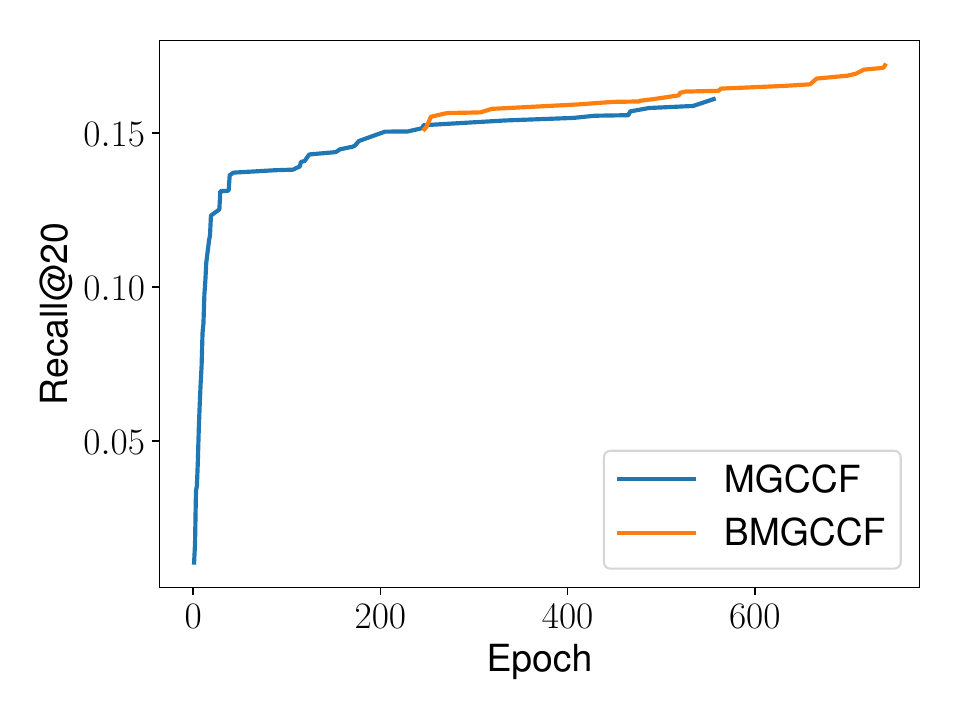}
  \caption{Training curve for MGCCF vs. BMGCCF (Amazon - CD).}
  \label{fig:training_curve_ngcf_az_cd}

\end{figure}
The learning curve comparison for training the original model and the Bayesian version of the model is shown in Figure~\ref{fig:training_curve_ngcf_az_cd} for the Amazon CD
dataset. We can observe that with our proposed solution, the training converges much faster. The Bayesian training framework also allows us to avoid overfitting in this case.

Conventional recommendation training procedure, especially in the
implicit recommendation setting, treat all of the unobserved user-item interactions as negative feedback (demonstrating a lack of interest). Our proposed approach aims to learn which of these unobserved interactions are most likely to be false negatives. We analyze the overlap between the edges that we remove from the negative candidates set with the edges in the validation and test set. As shown in Table~\ref{tab:edge_overlap_test_set}, our proposed Bayesian formulation is able to remove a significant percentage of test and validation edges from the negative sample pool.

\begin{table}[ht!]
\setlength{\tabcolsep}{2pt}
\caption{Edge overlap of the inferred graph with the test set.}
\label{tab:edge_overlap_test_set}
\begin{tabular}{@{}ccccl@{}}
\midrule[0.25ex]
          & \textbf{Am. CDs} &\textbf{Am. Books} & \textbf{Yelp2018} & \textbf{ML100Ks}\\ \midrule[0.25ex]
\textbf{BMGCF}                                        &     20.6\%       &        17.9\%                                                                                     &  13.6\%  &        12.3\%   \\ 
\textbf{BNGCF}                                                &   23.4\% &    30.0\%                             & 13.3\%  &     62.1\%         \\ \bottomrule
\end{tabular}

\end{table}
\vspace{-1em}
\section{CONCLUSION}
In this paper, we propose the use of non-parametric modelling and inference of graphs for various learning tasks. In the proposed model, a higher edge weight between two nodes is more likely if the nodes are close in terms of a distance metric. An appropriate distance metric can be chosen depending on the learning task which results in flexible, task-specific design of learning algorithms.
The proposed model is adapted to a Bayesian learning framework which aims to account for graph uncertainty. Experimental results demonstrate that the model can learn useful graphs that improve performance significantly over baseline algorithms for node classification, link prediction, and recommendation.

\bibliographystyle{apalike}
{
\bibliography{references}
}

\section{SUPPLEMENTARY MATERIAL}
\subsection{DESCRIPTION OF THE DATASETS}
For the semi-supervised node classification and link prediction tasks, we conduct experiments on benchmark citation network datasets (Cora~\cite{sen2008}, Citeseer~\cite{sen2008}, and Pubmed~\cite{namata2012}). In these datasets each node represents a research article and the undirected edges are formed according to citation links. Each node has a sparse
bag-of-words feature vector derived from the keywords of the document. The node labels indicate the primary research topics addressed in the articles. The statistics of the citation datasets are summarized in Table~\ref{table:dataset_statistics}.

\begin{table*}[htbp]
\setlength{\tabcolsep}{5pt}
\centering
\caption{Statistics of evaluation datasets for node classification and generative graph models. }
\vspace{0.1cm}
\begin{tabular}{lccccc}
\toprule
\textbf{Dataset} & \textbf{\# Classes} & \textbf{\# Features} & \textbf{\# Nodes} & \textbf{\# Edges} & \textbf{Edge Density} \\ \midrule
\textbf{Cora} & 7 & 1,433 & 2,485 & 5,069 & 0.04\% \\ 
\textbf{Citeseer} & 6 & 3,703 & 2110 &  3,668   & 0.04\% \\ 
\textbf{Pubmed} & 3 & 500 & 19,717 &  44,324   & 0.01\%  \\ 
\bottomrule
\end{tabular}
\label{table:dataset_statistics}
\end{table*}

For the experiments on the recommendation systems, we use four real-world and publicly available datasets: ML100K, Amazon-Books, Amazon-CDs and Yelp2018. For each dataset, we apply a pre-processing step to ensure that each user/item node in the dataset has sufficient interactions. We consider two threshold values $th_1$ and $th_2$, and filter out those users and those items with fewer than $th_1$ and $th_2$ interactions, respectively. The number of users and items after the preprocessing step are denoted by $|\mathcal{U}|$ and $|\mathcal{I}|$ respectively. The statistics of the preprocessed datasets are provided in Table~\ref{recstat}.

\begin{table*}[htbp]
\setlength{\tabcolsep}{5pt}
\renewcommand{\arraystretch}{1}
\centering
\caption{Statistics of evaluation datasets for recommender systems.}
\begin{tabular}{lcccccc}
\hline
\textbf{Dataset} & $\boldsymbol{(th_1, th_2)}$ & \textbf{($|\mathcal{U}|$, $|\mathcal{I}|$)}& \textbf{\# Edges} & \textbf{Edge Density}\\
\hline
\textbf{ML100K}
& (10, 10)
& (897, 823)
& 36330
& 0.0716
\\

\textbf{CDs}
& (50, 20)
& (2217, 2034)
& 68293
& 0.0151
\\

\textbf{Books}
& (250, 22)
& (1608, 2270)
& 69782
& 0.0191
\\

\textbf{Yelp2018}
& (22, 35)
& (2799, 1673)
& 119514
& 0.0255
\\
\bottomrule

\end{tabular}
\label{recstat}
\end{table*}

\subsection{GRAPH INFERENCE IN NON-PARAMETRIC MODEL}
For completeness, we provide a brief summary of the graph inference task considered in the main paper. We solve for the maximizer $\widehat{\mg}$ of the posterior distribution of the `true' graph $\mg$ as follows:
\begin{align}
 \widehat{\mg} &=
 \argmax_{\mg} p(\mg|\mg_{obs}, \md)\,,\label{app:opt:graph_inference}
\end{align}
Here, $\mg_{obs}$ denotes the observed graph and $\md$ is additional information. In our non-parametric model, this is equivalent to solving the following optimization problem for the adjacency matrix $\mathbf{A}_{\widehat{\mg}}$ of $\widehat{\mg}$: 
\begin{align}
      \label{app:opt:adj_inference}
  \mathbf{A}_{\widehat{\mg}}&= \argmin_{\substack{\mathbf{A}_{\mg} \in \mathbf{R_+}^{N\times N}, \\ \mathbf{A}_{\mg}=\mathbf{A}_{\mg}^{\top} }}
  \begin{aligned}[t]
  &\|\mathbf{A}_{\mg} \circ \mathbf{D}\|_{1,1} -\alpha\mathbf{1}^{\top}\log(\mathbf{A}_{\mg}\mathbf{1}) \\
  & \quad \quad \quad \quad \quad \quad \quad + \beta\|\mathbf{A}_{\mg}\|_F^2 \,.
 \end{aligned}
 \end{align}
We recall that $\mathbf{D}(\mg_{obs}, \md) \geq \mathbf{0}$ is a symmetric pairwise distance matrix which measures the dissimilarity between the nodes. $\alpha$ and $\beta$ are the hyperparameters of the prior distribution of the random graph $\mg$. 

\subsection{BAYESIAN GRAPH NEURAL NETWORK}
This section summarizes the novel Bayesian Graph Neural Network algorithm and provides some additional results for the node classification task.

\subsubsection{BGCN Algorithm}
\begin{algorithm}[H]
\caption{Bayesian GCN using non-parametric graph learning}
\label{alg:bgcn_non_param}
\begin{algorithmic}[1]
\STATE {\bfseries Input:}  $\mg_{obs}$, $\BX$, $\mathbf{Y_{\mathcal{L}}}$
\STATE {\bfseries Output:}  $p(\BZ|\mathbf{Y_{\mathcal{L}}},\BX,\mg_{obs})$

\STATE Train a node embedding algorithm using $\mg_{obs}$ and $\BX$ to obtain $\boldsymbol{z}_i$ for $1 \leq i \leq N$. Compute $\mathbf{D}_1$ using~\eqref{app:eq:dist1}. 

\STATE Train a base classifier using $\mg_{obs}$, $\BX$ and  $\mathbf{Y_{\mathcal{L}}}$ to obtain $\hat{c}_i$ for $1 \leq i \leq N$. Compute $\mathbf{D}_2$ using~\eqref{app:eq:dist2}. 

\STATE Compute $\mathbf{D}$ using~\eqref{app:eq:distance}.

\STATE  Solve the optimization problem in~\eqref{app:opt:adj_inference} to obtain $A_{\widehat{\mg}}$ (equivalently, $\widehat{\mg}$).

\FOR{$s=1$ {\bfseries to} $S$}
\STATE Sample weights $\mathbf{W}_{s}$ using MC dropout by training a GCN over the graph $\widehat{\mg}$.
\ENDFOR

\STATE Approximate $p(\BZ|\mathbf{Y_{\mathcal{L}}},\BX,\mg_{obs})$ using~\eqref{app:eq:MC_posterior}.
\end{algorithmic}
\end{algorithm}

We recall the notations from the main paper: $\BX$ is the feature matrix of the nodes in the observed graph $\mg_{obs}$ and $\mathbf{Y_{\mathcal{L}}}$ is the set of known training labels. $\mathbf{W}_{s}$ denotes the $s$-th sample of GCN weights trained on the inferred graph $\widehat{\mg}$ from the non-parametric model. $\boldsymbol{z}_i$ represents any suitable embedding of node $i$ and $\hat{c}_i \in \{1, 2, ... K\}$ is the obtained label of $i$-th node from a baseline node classification algorithm. $\mathcal{N}_i = \{j | (i,j) \in \mathcal{E}_{\mg_{obs}}\} \cup\{i\}$ is the neighbourhood of $i$-th node in $\mg_{obs}$. The distance matrix $\mathbf{D}$ is defined as follows:
\begin{align}
  \mathbf{D}(\BX,\mathbf{Y_{\mathcal{L}}},\mg_{obs}) &= \mathbf{D}_{1} (\BX,\mg_{obs}) + \delta \mathbf{D}_{2}(\BX,\mathbf{Y_{\mathcal{L}}},\mg_{obs})\,.\label{app:eq:distance} \\
  D_{1,ij} (\BX,\mg_{obs}) &= \|\boldsymbol{z}_i - \boldsymbol{z}_j\|^2\,,\label{app:eq:dist1}\\
     D_{2,ij}(\BX,\mathbf{Y_{\mathcal{L}}},\mg_{obs}) &= \frac{1}{|\mathcal{N}_i||\mathcal{N}_j|}\displaystyle{\sum_{k \in \mathcal{N}_i}\sum_{l \in \mathcal{N}_j}}\mathbb{1}_{(\hat{c}_k \neq \hat{c}_l)}\,.\label{app:eq:dist2}
\end{align}
Here, $\delta$ is a hyperparameter which controls the importance of $\mathbf{D}_2$ relative to $\mathbf{D}_1$. We need to compute the Monte Carlo approximation of the posterior distribution of labels, which is given as:
\begin{align}
p(\BZ|\mathbf{Y_{\mathcal{L}}},\BX,\mg_{obs}) \approx 
\dfrac{1}{S} \sum_{s=1}^S
p(\BZ|\mathbf{W}_{s},\mg_{obs},\BX)\,.
\label{app:eq:MC_posterior}
\end{align}
Pseudocode for the proposed BGCN algorithm is provided in Algorithm~\ref{alg:bgcn_non_param}.

\subsubsection{Results on fixed training-test split from~\cite{zhang2019}}
\begin{table}[htbp]
	\centering
	\caption{Classification accuracy (in $\%$) for Cora dataset.}
	\label{tab:cora_fixed}
	\footnotesize{
		\begin{tabular}{lcccc}
			\toprule[0.25ex]
		
			\textbf{Fixed split}  &\textbf{5 labels}        & \textbf{10 labels}         & \textbf{20 labels} \\
			\midrule
			\textbf{ChebyNet}             &67.9$\pm$3.1            &72.7$\pm$2.4             &80.4$\pm$0.7 \\
			
			\textbf{GCN}             &74.4$\pm$0.8            &74.9$\pm$0.7              & 81.6$\pm$0.5       \\
			\textbf{GAT}              &73.5$\pm$2.2            &74.5$\pm$1.3              &81.6$\pm$0.9  \\
		
			\textbf{BGCN}     &75.3$\pm$0.8   &76.6$\pm$0.8    &81.2$\pm$0.8     \\
			 \textbf{SBM-GCN}    &   59.3$\pm$1.3  &  \textbf{77.3$\pm$1.2}  &  \textbf{82.2$\pm$0.8} \\
		\textbf{BGCN (ours)}     &\textbf{76.0$\pm$1.1}   &76.8$\pm$0.9  &80.3$\pm$0.6    \\
			\bottomrule[0.25ex]
		\end{tabular}
	}
\end{table}
\begin{table}[htbp]
	\centering	
\caption{Classification accuracy (in $\%$) for Citeseer dataset.}
	\label{tab:citeseer_fixed}
		\footnotesize{
		\begin{tabular}{lcccc}
			\toprule[0.25ex]
			\textbf{Fixed split}  &\textbf{5 labels}        & \textbf{10 labels}         & \textbf{20 labels} \\ 
		
			\midrule
			\textbf{ChebyNet}                     &53.0$\pm$1.9            &67.7$\pm$1.2              &70.2$\pm$0.9 \\
		
			\textbf{GCN}             &55.4$\pm$1.1            &65.8$\pm$1.1              &70.8$\pm$0.7       \\
			
			\textbf{GAT}        &55.4$\pm$2.6            &66.1$\pm$1.7              &70.8$\pm$1.0 \\
		
			\textbf{BGCN}   &57.3$\pm$0.8   &70.8$\pm$0.6    &72.2$\pm$0.6\\
			\textbf{SBM-GCN}    &   20.8$\pm$2.0  &  66.3$\pm$0.6  & 71.7$\pm$0.1  \\
			\textbf{BGCN (ours)}        &\textbf{59.0$\pm$1.5}   &\textbf{71.7$\pm$0.8}     &\textbf{72.6$\pm$0.6} \\
			\bottomrule[0.25ex]
		\end{tabular}
	}
	\end{table}
\begin{table}[htbp]
	\centering
\caption{Classification accuracy (in $\%$) for Pubmed dataset.}
 \label{tab:pubmed_fixed}
	\footnotesize{
		\begin{tabular}{lcccc}
			\toprule[0.25ex]
			\textbf{Fixed split}  &\textbf{5 labels}        &\textbf{10 labels}         &\textbf{20 labels} \\ 
		
			\midrule
			\textbf{ChebyNet}                    &68.1$\pm$2.5            &69.4$\pm$1.6              &76.0$\pm$1.2             \\
	
			\textbf{GCN}             &69.7$\pm$0.5            &72.8$\pm$0.5            &78.9$\pm$0.3  \\
			\textbf{GAT}   &70.0$\pm$0.6   & 71.6$\pm$0.9            &76.9$\pm$0.5 \\
		
		\textbf{BGCN}   &70.9$\pm$0.8           &72.3$\pm$0.8    &76.6$\pm$0.7\\
		\textbf{SBM-GCN}    &  64.8$\pm$0.8   &  71.7$\pm$0.7  &  \textbf{80.6$\pm$0.4 }\\
			\textbf{BGCN (ours)}       &\textbf{73.3$\pm$0.7}      &\textbf{73.9$\pm$0.9}    &79.2$\pm$0.5\\
			\bottomrule[0.25ex]
		\end{tabular}
	}
\end{table}
In the main paper, we consider random partitioning of the nodes in training and test sets and report the average accuracies across different splits. We conduct another experiment where the same fixed training-test split of~\cite{zhang2019} is used for 50 random intializations of the GCN weights. The classification results for this setting are provided in Tables~\ref{tab:cora_fixed},
\ref{tab:citeseer_fixed} and~\ref{tab:pubmed_fixed}.
\subsection{BAYESIAN VGAE ALGORITHM FOR LINK PREDICTION}
We recall that the function $\mathcal{J}(\mg, \mg_{obs})$ returns a graph such that
the unobserved entries of the adjacency matrix of $\mg_{obs}$ are replaced by the corresponding entries of $\mg$. The distance matrix for the non-parametric graph inference is defined as:
 \begin{align}
     D_{ij} (\BX,\mg_{obs}) &= \|\mathbb{E}_q[\boldsymbol{z}_i] - \mathbb{E}_q[\boldsymbol{z}_j]\|^2\,,\label{app:eq:dist_gae}
\end{align}
where, $q(\boldsymbol{Z}|\mg_{obs}, \BX)$ is the approximate posterior distribution of unsupervised node representations $\boldsymbol{Z}$ from a Variational Graph Auto-Encoder (VGAE) model. In the proposed Bayesian VGAE, the inference distribution is modelled as follows:
\begin{align}
q(\boldsymbol{Z}|\mg_{obs}, \BX) &= \int q(\boldsymbol{Z}|\mathcal{J}(\mg, \mg_{obs}), \BX) p(\mg|\mg_{obs}, \BX) d \mg\,,\nonumber\\
&\approx q(\boldsymbol{Z}|\mathcal{J}(\widehat{\mg}, \mg_{obs}), \BX)\,.\label{app:eq:approx_q}
\end{align}
Here $\widehat{\mg}$ is the inferred graph from the non-parametric model. The resulting algorithm is summarized in Algorithm~\ref{alg:bgae_non_param}
\begin{algorithm}[H]
\caption{Bayesian VGAE}
\label{alg:bgae_non_param}
\begin{algorithmic}[1]
\STATE {\bfseries Input:}  $\mg_{obs}$, $\BX$
\STATE {\bfseries Output:}  $q(\boldsymbol{Z}|\mg_{obs}, \BX)$

\STATE Train a node embedding algorithm using $\mg_{obs}$ and $\BX$ to obtain $\boldsymbol{z}_i$ for $1 \leq i \leq N$.  

\STATE Compute $\mathbf{D}$ using eq.~\eqref{app:eq:dist_gae}.

\STATE  Solve the optimization problem in~\eqref{app:opt:adj_inference} to obtain $A_{\widehat{\mg}}$ (equivalently, $\widehat{\mg}$).

\STATE Build a new graph $\mathcal{J}(\widehat{\mg}, \mg_{obs})$ and train the auto-encoder on it to obtain $q(\boldsymbol{Z}|\mg_{obs}, \BX)$ (eq.~\eqref{app:eq:approx_q}).
\end{algorithmic}
\end{algorithm}

\subsection{RECOMMENDATION SYSTEMS}
\subsubsection{Algorithm}
We recall that $\{>_u\}_{train}$ is the set of training ranknings in observed graph $\mg_{obs}$. The distance between the $u$-th user and the $i$-th item is defined as:
\begin{align}
D_{u,i}(\{>_{\mathcal{U}}\}_{train}, \mg_{obs}) = 1 - \frac{e_u \cdot e_i}{\lvert \lvert e_u \rvert \rvert_2 \lvert \lvert e_i \rvert \rvert_2 }\,.\label{app:eq:dist_cosine}
\end{align}
Here, $e_u$ and $e_i$ are the representations from the base node embedding algorithm for the $u$-th user and the $i$-th item, respectively. We define $\widetilde{G} = \mathcal{J}_r(\mg, \mg_{obs})$ by removing a fraction of links with the highest edge weights in $\mg$  from the negative pool of interactions according to $\mg_{obs}$. In the Bayesian version of the recommendation system, we need to compute:
\begin{align}
&p(\{>_{\mathcal{U}}\}_{test}|\{>_{\mathcal{U}}\}_{train},\mg_{obs}) = \int p(\{>_{\mathcal{U}}\}_{test}|\mg_{obs}, \mathbf{W})\,\nonumber\\
& \quad p( \mathbf{W}|\{>_{\mathcal{U}}\}_{train}, \widetilde{\mg}) p(\mg| \mg_{obs}, \{>_{\mathcal{U}}\}_{train})\,d\mg\,d \mathbf{W} \,, \nonumber\\
&\approx p(\{>_{\mathcal{U}}\}_{test}|\mg_{obs}, \widehat{\mathbf{W}})\,.\label{app:eq:mc"_bbpr}
\end{align}
We perform non-parametric graph inference to obtain $\widehat{\mg}$, then compute $\widetilde{G} = \mathcal{J}_r(\widehat{\mg}, \mg_{obs})$ and minimize the BPR loss to form the estimate of the weights:
\begin{align}
    \widehat{\mathbf{W}} = \argmax_{\mathbf{W}} \Hquad p( \mathbf{W}|\{>_u\}_{train}, \widetilde{G}) \,.\label{app:eq:weight_opt}
\end{align}
The resulting algorithm is summarized in Algorithm~\ref{alg:rec_non_param}.
\begin{algorithm}[ht!]
\caption{BPR loss minimization with non-parametric graph learning}
\label{alg:rec_non_param}
\begin{algorithmic}[1]
\STATE {\bfseries Input:}  $\{>_u\}_{train}$, $\mg_{obs}$
\STATE {\bfseries Output:}  $\{>_u\}_{test}$

\STATE Train a base node embedding algorithm using $\mg_{obs}$ and $\{>_{\mathcal{U}}\}_{train}$ to obtain embeddings for all users and items.

\STATE Compute $\mathbf{D}$ as the pairwise cosine distance between the embeddings (eq.~\eqref{app:eq:dist_cosine}).

\STATE  Solve the optimization problem in~\eqref{app:opt:adj_inference} using $\mathbf{D}$ to obtain $\widehat{\mg}$.

\STATE Form the modified graph $\widetilde{\mg} = \mathcal{J}_r(\widehat{\mg}, \mg_{obs})$. 

\STATE Minimize BPR loss with negative pool according to $\widetilde{\mg}$ to obtain $\mathbf{\widehat{W}}$ (eq.~\eqref{app:eq:weight_opt}).

\STATE Obtain the test set rankings using the embeddings to form $\{>_{\mathcal{U}}\}_{test}$ (eq.~\eqref{app:eq:mc"_bbpr})
\end{algorithmic}
\end{algorithm}

\subsubsection{Definitions of the performance evaluation metrics}

 \begin{itemize}
    
\item  \textit{Recall@k} denotes the proportion of the true (preferred) items from the top-$k$ recommendation. For a user $u\in\mathcal{U}$, the algorithm recommends an ordered set of top-$k$ items $I_k(u)\in\mathcal{I}$. There is a set of true preferred items for a user $\mathcal{I}_u^+$ and a number of true positive $|\mathcal{I}_u^+\cap I_k(u)|$. The recall of a user is defined as follows : 
  \[\mathrm{Recall@k}=\frac{|\mathcal{I}_u^+\cap I_k(u)|}{|\mathcal{I}_u^+|}\]
      
   \item  \textit{NDCG@k}: Normalized Discounted Cumulative Gain (NDCG)~\cite{Jrvelin2000IREM} computes a score for $I(u)$ which emphasizes higher-ranked true positives.  $D_k(n)=(2^{rel_{n}}-1)/\log_2(n+1)$ accounts for a relevancy score
      $rel_n$. We consider
      binary responses, so we use a binary relevance score:
      $\mathrm{rel}_n=1$ if $i_n\in\mathcal{I}_u^+$ and 0 otherwise.
      \[\mathrm{NDCG}@k=\frac{\mathrm{DCG_k}}{\mathrm{IDCG_k}}=\frac{\sum_{n=1}^{|\mathcal{I}|}\mathrm{D}_k(n)[i_n\in\mathcal{I}_u^+]}{\sum_{n=1}^{|I_k(u)|}D_k(n)}\]
 \end{itemize}
 
\end{document}